\pdfoutput=1

\documentclass[11pt]{article}

\usepackage[]{acl}

\usepackage{times}
\usepackage{latexsym}
\usepackage{MnSymbol}
\usepackage{amsmath}
\usepackage{expex}

\usepackage[T1]{fontenc}

\usepackage[utf8]{inputenc}

\usepackage{microtype}
\usepackage{booktabs}

\usepackage{inconsolata}
\usepackage{graphicx}

\usepackage{dblfloatfix}
\usepackage[skins]{tcolorbox}
\newtcolorbox{myframe}[2][]{%
  enhanced,colback=yellow!20!white,colframe=black,coltitle=black,boxrule=0.5pt,
  fonttitle=\bf,
  attach boxed title to top left={yshift=-0.3\baselineskip-0.4pt,xshift=2mm},
  boxed title style={tile,size=minimal,left=0.5mm,right=0.5mm,
    colback=white,before upper=\strut},
  title=#2,#1
}

\usepackage{ifthen}
\newboolean{includeappendix}   

\setboolean{includeappendix}{true}

\newcommand{\haveappendix}[1]
    {\ifthenelse{\boolean{includeappendix}}{#1}{}}

\title{LIEDER: Linguistically-Informed Evaluation for Discourse Entity Recognition}

\author{Xiaomeng Zhu \\
  Department of Linguistics \\
  Yale University  \\
  \texttt{miranda.zhu@yale.edu} \\\And
  Robert Frank \\
  Department of Linguistics \\
  Yale University \\
  \texttt{bob.frank@yale.edu} \\ }

\begin{document}
\maketitle

\begin{abstract}
Discourse Entity (DE) recognition is the task of identifying novel and known entities introduced within a text. While previous work has found that large language models have basic, if imperfect, DE recognition abilities \citep{schuster-linzen-2022-sentence}, it remains largely unassessed which of the fundamental semantic properties that govern the introduction and subsequent reference to DEs they have knowledge of. We propose the Linguistically-Informed Evaluation for Discourse Entity Recognition (LIEDER) dataset that allows for a detailed examination of language models' knowledge of four crucial semantic properties: \textsc{existence}, \textsc{uniqueness}, \textsc{plurality}, and \textsc{novelty}. We find evidence that state-of-the-art large language models exhibit sensitivity to all of these properties except \textsc{novelty}, which demonstrates that they have yet to reach human-level language understanding abilities. 

\end{abstract}

\section{Introduction}
One central component of language understanding is the ability to recognize entities in text. A large body of research in Natural Language Processing focuses on the task of Named Entity Recognition, where a system must identify whether a noun phrase (NP), typically a proper name, refers to a known individual of a certain semantic class \citep{nersurvey}. The recognition of discourse entities (DEs), in contrast, involves identifying not only the occurrence of known entities but also novel ones that are introduced within a text. The recognition of DEs takes place at two sites: introduction sites and reference sites. Introduction refers to the first time where an entity appears in a discourse.  Reference sites are subsequent mentions of an entity that has been previously introduced. 

As humans, not only are we able to recognize DEs at both of these sites, but we also have knowledge of \textbf{how} to coordinate the introduction and subsequent reference to entities using appropriate linguistic means. For example, we know that the introduction of DEs is typically done using indefinite NPs such as \textit{a man} in `\textit{A man walked into the room}.'
We also know that subsequent mentions often involve definite NPs like \textit{the man} in `\textit{The man sat down}.' 

DE recognition is an important component of more complex semantic understanding tasks such as coreference resolution. Coreference relationships cannot be established between entities that have not been introduced into the discourse.
\pex[aboveexskip=3pt, belowexskip=3pt]<crex>
\a John owns a dog. The dog is cute.
\a John doesn't own a dog. \#The dog is cute.
\xe
For example, in (\getref{crex}a), the NP \textit{the dog} in the second sentence and \textit{a dog} in the first sentence refer to the same entity. This is not the case in (\getref{crex}b) because no entities have been introduced in the first sentence, which makes the continuation in (\getref{crex}b) infelicitous. Therefore, before establishing coreference relationships, language models first need to perform DE recognition, 

\citet{schuster-linzen-2022-sentence} present an evaluation suite for DE recognition that focuses on the question of whether language models are sensitive to the linguistic context in which DEs are introduced. They find that transformer-based language models do not always demonstrate a clear preference for referring to entities that have been properly introduced into the discourse. While this work provides important insight into LM abilities with discourse reference, it does not engage directly with the underlying linguistic properties responsible for DE introduction and reference. As a result, it does not provide a means of assessing more precisely what LMs know about the linguistic encoding of discourse reference. 

Semantics research has established properties of definite and indefinite NPs from which their use in introducing and referring to entities follows, four of which are particularly relevant here: \textsc{existence}, \textsc{uniqueness}, \textsc{plurality}, and \textsc{novelty}. We will define and discuss these properties in detail in Section \ref{sec:background}. A good language model (LM) should reflect knowledge of all of these properties. In this paper, we provide a novel dataset, which builds on Schuster and Linzen's work, that provides a method of testing these properties directly.\footnote{All code, data, and results are available at \url{https://github.com/xiaomeng-zhu/LIEDER}.} Our results, across a number of state-of-the-art (SOTA) large language models (LLMs), provide evidence for knowledge of \textsc{existence}, \textsc{uniqueness}, and \textsc{plurality} (all conditions on the use of definite NPs to refer to DEs), but difficulty with \textsc{novelty} (a condition on the introduction of DEs by indefinite NPs) unless information about distinctiveness is made explicit. In addition, we find that transformer LMs, unlike humans, show strong sensitivity to linear distance in establishing DE reference.  Taken together, these results suggest that SOTA LLMs do not reach human-level language understanding abilities.

\section{Assessing Discourse Entity Recognition} \label{sec:originalpaper}

\citet{schuster-linzen-2022-sentence} (henceforth SL) develop an evaluation suite that probes LLM performance on DE recognition. Such evaluation was centered on the ability of indefinite phrases to introduce DEs that can be referred to by subsequent occurrences of definites.  They identified pairs of contexts that differ in their ability to support DE introduction.\haveappendix{\footnote{For a review of the pairs of contexts introduced by \citet{schuster-linzen-2022-sentence}, see Appendix \ref{subsec:addplots1}.}}
The simplest case, and the one we focus on in our experiments, is  \textit{affirmative-negation}: indefinites in the object position of affirmative sentences introduce DEs that can be referred to by a following definite NP, but they do not in the object of sentences with negation. This is seen in the following example, where we use F to represent a felicitous completion, and I to represent an infelicitous completion:

\pex[aboveexskip=3pt, belowexskip=3pt]<metric1>
\textsc{Context}: John owns a dog but he doesn't own a cat.
\a F: His dog follows him everywhere.
\a I: His cat follows him everywhere.
\xe

SL argue that if a language model is sensitive to the difference between contexts which do and do not introduce DEs (such as the two conjuncts in the first sentence in (\getref{metric1})
), we would expect the following inequality to hold for the probabilities assigned to felicitous (F) and infelicitous (I) continuations:
\setlength{\belowdisplayskip}{3pt} \setlength{\belowdisplayshortskip}{3pt}
\setlength{\abovedisplayskip}{3pt} \setlength{\abovedisplayshortskip}{3pt}
\begin{equation*}
p(\textsc{F}|\textsc{Context})>p(\textsc{I}|\textsc{Context})    
\end{equation*}
They found that the models they examined (which included GPT-2 variants and GPT-3) showed above chance performance on distinguishing F and I continuations in the context of \textit{affirmative-negation}, with  GPT-3 showing the most human-like performance. 

SL further explored the systematicity of these contrasts,  where a contrast is counted as systematic only when the model correctly predicts felicity on all possible orderings of the operators and nouns in the conjoined clauses that comprise the context: 
\pex[aboveexskip=3pt, belowexskip=3pt, interpartskip=3pt]<distance>
\a Bob owns a dog but he doesn't own a cat.
\a Bob owns a cat but he doesn't own a dog.
\a Bob doesn't own a cat but he owns a dog.
\a Bob doesn't own a dog but he owns a cat.
\xe

The relative order of the affirmative and negative sentences should not impact the felicity of subsequent definite descriptions. 
The continuation \textit{His dog follows him everywhere} is felicitous after either (\getref{distance}a) or (\getref{distance}c). Similarly, the continuation \textit{His cat follows him everywhere} remains felicitous after  (\getref{distance}b) or (\getref{distance}d). 
However, SL find that the models are much less successful under this more demanding measure; even GPT-3 systematically distinguishes felicitous and infelicitous continuations only barely above 50\% of the time, with other models showing lower performance. Schuster and Linzen do not, however, identify what underlies the models' failure in systematic performance on this task. 

While intriguing, these results do not tell us what specifically causes difficulty for LLMs in this task, and how it relates to semantic properties of definite and indefinite NPs. To evaluate language models' DE recognition abilities, we believe that it is important to decompose models' performance in a more granular manner.  We turn now to a paradigm that builds on this previous work to do precisely this.

\section{Criteria for Discourse Entity Recognition} \label{sec:background}
Informed by theoretical research in natural language semantics, we propose that a thorough evaluation of DE recognition and reference abilities should examine language models' knowledge of four fundamental semantic properties: \textsc{existence}, \textsc{uniqueness}, \textsc{plurality}, and \textsc{novelty}. We will define and discuss each of them in order.

\paragraph{Existence}
As SL argued, a language model with human-level understanding abilities should only use definite descriptions to refer to entities that have been introduced into the discourse. We define this requirement as \textsc{existence} \citep{russell1905denoting}.
For example, given the context \textit{John doesn't own a dog}, a language model should recognize that the continuation \textit{The dog barks at night} is infelicitous because of the non-existence of a dog DE.

\paragraph{Uniqueness}
A language model should use a singular definite description to refer to a previously introduced entity only when the referent is unique relative to the discourse. We will call this requirement \textsc{uniqueness} \citep{russell1905denoting,heim1998semantics}. For example, given the context \textit{John owns a dog and Mark owns a dog too}, the model should consider the continuation \textit{The dog barks at night} as infelicitous. Since more than one dog has been introduced into the discourse, the singular definite phrase is not licensed. 
On the other hand, if the context is  \textit{John does not own a dog but Mark owns a dog}, \textsc{uniqueness} is satisfied, so the same continuation should be judged to be felicitous.

\paragraph{Plurality}
A language model should use a plural definite description only if the set of DEs contains more than one individual of the relevant sort, a requirement we call \textsc{plurality} 
\citep{landman1989groups}. Notice that \textsc{uniqueness} and \textsc{plurality} cannot be satisfied or violated at the same time -- referent expressions that require uniqueness (i.e., singular definites) are infelicitous after contexts that satisfy \textsc{plurality}. In contrast, referent expressions that require plurality (i.e., plural definites) are infelicitous after contexts that satisfy \textsc{uniqueness}. 
For example, the context \textit{John owns a dog and Mark owns a dog too} satisfies \textsc{plurality} since there are two dog DEs, and therefore supports a plural but not singular continuation (i.e., \textit{The dogs bark/*dog barks}).

\paragraph{Novelty}
The last requirement concerns the use of indefinite NPs: a language model should recognize that an occurrence of an indefinite noun phrase is associated with the introduction of a new entity into the discourse. Following \citet{heim1982semantics}, we will call this requirement \textsc{novelty}. In the context sentence \textit{John owns a dog and Mark owns a dog too}, this means that two distinct dogs are introduced as DEs. 

Table \ref{tab:property} summarizes relevant expression types and the corresponding requirements that a language model should know in order to correctly introduce and refer to DEs. In the next section, we will describe our evaluation dataset and show how it evaluates model performance with respect to the four requirements.

\begin{table}  
\begin{tabular}{cc}
\toprule
Expression &  Requirements\\
\midrule
Indefinites & \textsc{novelty}\\
Singular definites &  \textsc{existence}, \textsc{uniqueness}\\
Plural definites &  \textsc{existence}, \textsc{plurality}\\
\bottomrule
\end{tabular}
\caption{Expressions used for introducing and referencing DEs and their corresponding requirements.}
\label{tab:property}
\end{table}

\section{The LIEDER Dataset} \label{sec:dataset}

The Linguistically-Informed Evaluation of DE Recognition (LIEDER) evaluation dataset adapts the structure of SL's paradigm: a context example is provided that consists of two conjoined clauses, each containing an indefinite NP with the same head noun. This is followed by a continuation, test sentence containing a definite description. As in SL, we vary the conjoined clauses as to whether they introduce DEs or not (affirmative or negative).  However, in order to evaluate the four linguistic properties defined in the previous section, we make two innovations: (i) we allow zero, one, or both of the conjoined clauses to include negation and thereby fail to introduce a DE; and (ii) we allow the definite description in the continuation to be either singular or plural. Example items in our dataset are shown in Table \ref{tab:newdatasing}.
\begin{table*}[t] \resizebox{\textwidth}{!}{
\centering
\begin{tabular}{llll}
\toprule
Context type & Context & Singular Continuation & Plural Continuation\\
  \midrule
 \texttt{pos\_neg} & John owns a dog but Mark doesn't own a dog.  &The dog is very cute. &\#The dogs are very cute.\\
\texttt{neg\_pos} &  John doesn't own a dog but Mark owns a dog.  & The dog is very cute.&\#The dogs are very cute.\\
\texttt{pos\_pos} & John owns a dog and Mark owns a dog too. & \#The dog is very cute.&The dogs are very cute.\\
\texttt{neg\_neg}& John doesn't own a dog and Mark doesn't own a dog either. &\#The dog is very cute.&\#The dogs are very cute.\\
\bottomrule
\end{tabular}}
\caption{Example contexts and continuations in the LIEDER dataset. Infelicitous continuations are marked with \#. }
\label{tab:newdatasing}
\end{table*}

The \textit{Context type} column indicates which sides of the conjunction introduce DEs. For example, \texttt{pos\_neg} indicates that a DE is introduced in the first conjunct (\texttt{pos}), and no DEs are introduced in the second conjunct (\texttt{neg}). Since we consider all four context types with both singular and plural continuations, there are 8 different context-continuation combinations. These 8 combinations are then crossed with 16 distinct base sentence pairs for the context and continuation, resulting in 128 examples in total.\footnote{SL consider other pairs of sentential operators that differ in whether they introduce DEs, specifically \textit{managed-failed} and \textit{know-doubt}. The LIEDER dataset also includes contexts that combine these sentential operators, which sums to 384 examples in total.
\haveappendix{See the Appendix for results regarding these other operator contrasts.}
} 

Different combinations of context type and continuation result in differences in felicity of our two-sentence discourse. The singular continuation is felicitous only when the context type is either \texttt{pos\_neg} or \texttt{neg\_pos} since these contexts introduce exactly one DE. In contrast, the plural continuation is felicitous only when the type is \texttt{pos\_pos}. This means that of the eight context-continuation pairs, three are felicitous and five are infelicitous. 

Success in the LIEDER dataset requires that a model accurately distinguish felicitous from infelicitous context-continuation pairs. 
Because metalinguistic judgments elicited from language models may not reflect the full extent of the model's knowledge \citep{hu2023prompting}, we instead compare felicity using the probabilities the model assigns to the continuation given the context.  We assume that the probability a model assigns to a felicitous case should be greater than the probability it assigns to an infelicitous one. 
With 3 felicitous pairs and 5 infelicitous ones, this means we have 15 informative probability comparisons in total. These are depicted in Table \ref{tab:comparison}.  

Importantly, success in each of these comparisons can be tied to the linguistic requirements described in Section \ref{sec:background} that are involved in the introduction of and reference to DEs.  For example, if the model assigns higher probability to a continuation with a singular definite in a \texttt{neg\_pos} context as compared to a \texttt{neg\_neg} context, this provides evidence for the model's awareness of the \textsc{existence} requirement that singular definite NP imposes; otherwise, the singular should be possible in this context. On the other hand, if the model assigns higher probability to a singular definite in a \texttt{pos\_neg} context as compared to a \texttt{pos\_pos} context, this indicates that the model is aware of the \textsc{uniqueness} requirement imposed on singular definiteness and the \textsc{novelty} condition imposed on indefinites, since otherwise the \texttt{pos\_pos} context could be taken to introduce only a single DE. In addition, if the model assigns higher probability to a singular definite in a \texttt{neg\_pos} context than it does to a plural definite in the same context, this provides evidence of sensitivity to the \textsc{plurality} requirement on plural definites, as the context introduces only a single entity. 
Because of the correspondences between the example types and the linguistic requirements, LIEDER can therefore be used to assess the details of the knowledge of DEs in a language model.

\begin{table}[] 
\centering
\scalebox{0.6}{
\begin{tabular}{ccc}
\toprule
Comparison Type                        & Requirement    & Section \\
\midrule
\texttt{p(sg|pos\_neg)>p(sg|pos\_pos)} & \textsc{uniqueness}, \textsc{novelty} &  \ref{subsubsec:sing}\\
\texttt{p(sg|neg\_pos)>p(sg|pos\_pos)} & \textsc{uniqueness}, \textsc{novelty} &  \ref{subsubsec:sing}\\
\texttt{p(sg|neg\_pos)>p(sg|neg\_neg)} & \textsc{existence}  &  \ref{subsubsec:sing}\\
\texttt{p(sg|pos\_neg)>p(sg|neg\_neg)} & \textsc{existence} &  \ref{subsubsec:sing}\\
\hline
\texttt{p(pl|pos\_pos)>p(pl|pos\_neg)} & \textsc{plurality} &  \ref{subsubsec:plu}\\
\texttt{p(pl|pos\_pos)>p(pl|neg\_pos)} & \textsc{plurality} &  \ref{subsubsec:plu}\\
\texttt{p(pl|pos\_pos)>p(pl|neg\_neg)} & \textsc{existence}, \textsc{plurality} &  \ref{subsubsec:plu}\\
\hline
\texttt{p(sg|pos\_neg)>p(pl|pos\_neg)} & \textsc{plurality} &  \ref{subsubsec:singandplur}\\
\texttt{p(sg|pos\_neg)>p(pl|neg\_pos)} & \textsc{plurality} &  \ref{subsubsec:singandplur}\\
\texttt{p(sg|pos\_neg)>p(pl|neg\_neg)} & \textsc{existence}, \textsc{plurality}  & \ref{subsubsec:singandplur}\\
\texttt{p(sg|neg\_pos)>p(pl|neg\_pos)} & \textsc{plurality} &  \ref{subsubsec:singandplur}\\
\texttt{p(sg|neg\_pos)>p(pl|pos\_neg)} & \textsc{plurality} &  \ref{subsubsec:singandplur}\\
\texttt{p(sg|neg\_pos)>p(pl|neg\_neg)} & \textsc{existence}, \textsc{plurality}  & \ref{subsubsec:singandplur}\\
\texttt{p(pl|pos\_pos)>p(sg|pos\_pos)} & \textsc{uniqueness}, \textsc{novelty} & \ref{subsubsec:singandplur}\\
\texttt{p(pl|pos\_pos)>p(sg|neg\_neg)} & \textsc{existence} & \ref{subsubsec:singandplur} \\
\bottomrule
\end{tabular}}
\caption{Comprehensive list of all comparison types, the requirements they test, and the experiment that tests for them.}
\label{tab:comparison}
\end{table}

\section{Experiment 1: Applying LIEDER} \label{sec:exp1}

\paragraph{Models}
We investigated the performance of five open-source (Llama 2-7B, 13B, and 70B \citep{touvron2023llama2}, Llama 3-8B and 70B \citep{llama3}) and two closed-source LLMs (\href{https://platform.openai.com/docs/models/gpt-base}{GPT \texttt{babbage-002} and \texttt{davinci-002}}) on LIEDER through the Huggingface transformer API \citep{wolf2019huggingface} and the OpenAI API respectively.
\footnote{The number of parameters in \texttt{babbage-002} and \texttt{davinci-002} are not publicly available. We were also not able to examine more recent LLMs released by OpenAI because the API for these models does not support access to the log probabilities of prompts.} 

\paragraph{Metric}
To perform the probability comparisons discussed in  Section \ref{sec:dataset}, we provided the model with the context of each test item and calculated the total log probability for the entire continuation.  For each comparison type,  we compare the log probability of a felicitous continuation with an infelicitous one, and judge the model as accurate if the felicitous probability is higher. We then calculate the accuracy for each comparison type over the 16 items of each type.

\paragraph{Human Judgments}
In addition to evaluating large language models with our dataset, we also conducted an experiment to elicit human judgments. Specifically, participants were asked to provide a rating on the acceptability of a continuation given a context using a continuous slider with possible values ranging from 1 to 7. Each participant provided ratings for 16 experimental sentences (one for each context type-continuation pair), each with different lexical content. This means that comparisons of items were done across subjects. Using the Prolific platform, we recruited 80 participants who were native English speakers with perfect or corrected vision and without language disorders. Acceptability ratings on the same sentences were averaged, and the resulting averages were compared within an item to compute accuracies that could be compared to the language model results. \haveappendix{See Appendix \ref{sec:morehuman} for our experimental interface and more details on methodology.}

\subsection{Results}
Because all continuations in LIEDER involve definite noun phrases, we refer to singular definite continuations simply as singular continuations and plural definites as plural continuations. Discussion of the results will be broken down based on the plurality of the continuations involved in the comparisons.

\subsubsection{Singular Continuations} \label{subsubsec:sing}
Singular continuations are felicitous after contexts where one and only one relevant DE is introduced (those that satisfy both \textsc{existence} and \textsc{uniqueness}), namely \texttt{pos\_neg} and \texttt{neg\_pos}. On the other hand, they are infelicitous for contexts that are of the type \texttt{pos\_pos} (violating \textsc{uniqueness}) and \texttt{neg\_neg} (violating \textsc{existence}). As a result, if the language models we are examining reach human-level understanding abilities, we would expect that the probabilities associated with the two felicitous cases are greater than the two infelicitous cases respectively, which gives us four comparison types.

\begin{figure*}[h] 
\centering
\includegraphics[scale=0.78]{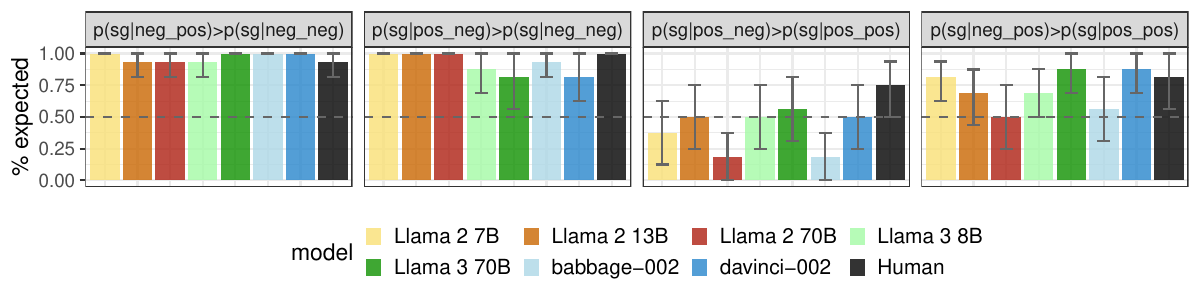}
\caption{Results for singular continuations by model and comparison type. The dotted lines indicate chance performance and the error bars indicate bootstrapped 95\% confidence intervals. 
}
\label{fig:exp1sing}
\end{figure*}

The results for these comparisons are shown in Figure \ref{fig:exp1sing}, respectively in the four panels. As the first two panels in the figure show, all models (and humans) have ceiling or near ceiling performance on dispreferring singular continuations that follow contexts where no DEs have been introduced (\texttt{neg\_neg)}. In other words, for a given singular continuation, they have a strong preference for contexts where one and only one DE is introduced over ones where none are introduced. Such preference indicates that all models know \textsc{existence}.

In contrast, in the last two panels in Figure \ref{fig:exp1sing}, we see that model accuracies are uniformly lower when the infelicitous context is \texttt{pos\_pos} as compared to \texttt{pos\_neg} or \texttt{neg\_pos} (i.e., two relevant DEs are introduced). 
This drop in accuracy suggests that the language models do not consider \texttt{pos\_pos} to be worse than \texttt{pos\_neg} or \texttt{neg\_pos} in licensing the same singular definite continuations.   
As a side note, a greater number of model parameters does not necessarily translate into higher performance. In fact, increasing parameter count in the Llama 2 series yields ever worse performance in panel 4. 

Why might \texttt{pos\_pos} contexts be confusing for large language models? With respect to the linguistic requirements on DE introduction and reference, there are two possible answers to this question:

\begin{myframe}{Hypothesis 1}
During training, the models have successfully learned the \textsc{existence} requirement, but they failed to learn \textsc{uniqueness}.
\end{myframe}

\begin{myframe}{Hypothesis 2}
During training, the models have successfully learned both the \textsc{existence} and \textsc{uniqueness} requirements but fail to recognize that two distinct DEs have been introduced in \texttt{pos\_pos} contexts, resulting in difficulties in distinguishing the infelicitous \texttt{pos\_pos} from felicitous \texttt{pos\_neg} and \texttt{neg\_pos}. To put it in another way, they fail at the \textsc{novelty} requirement.
\end{myframe}

At this stage, we lack evidence that supports one hypothesis over another. Experiment 2 focuses on teasing these two hypotheses apart.

\begin{figure}[h]
\centering
\includegraphics[scale=0.5]{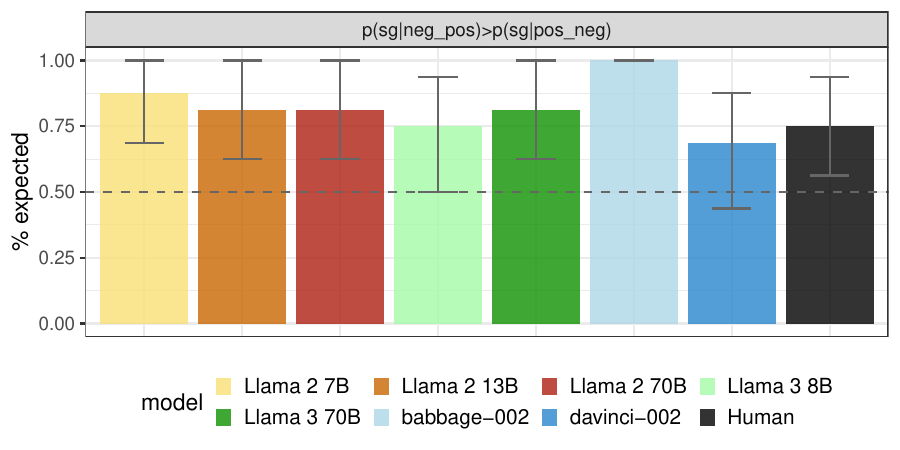}
\caption{Preference for \texttt{neg\_pos} over \texttt{pos\_neg} by model.}
\label{fig:exp1sing-f}
\end{figure}

From the last two panels in Figure \ref{fig:exp1sing}, we can see that accuracy of all LMs for \texttt{p(sg|neg\_pos)>p(sg|pos\_pos)} is uniformly higher than \texttt{p(sg|pos\_neg)>p(sg|pos\_pos)} across models, while human performance differs very little. 
We suggest that the source of this contrast is a preference for singular definites in the context of \texttt{neg\_pos} over \texttt{pos\_neg}, both of which ought to be felicitous contexts. 
Figure \ref{fig:exp1sing-f}  evalulates this claim, illustrating the percentage of test examples where models and humans consider \texttt{neg\_pos} to a better context for singular definite continuations than \texttt{pos\_neg}. All models exhibit this preference for over half of the examples.
Interestingly and perhaps surprisingly, humans also demonstrate a distance effect, showing a preference for \texttt{neg\_pos} over \texttt{pos\_neg} contexts for singular definites on a majority of examples.  Thus, the presence of \textsc{distance} sensitivity need not be interpreted to be a deficiency of LM performance, but perhaps is a reflection of patterns in human language use. Nonetheless, though humans show a \textsc{distance} effect, they demonstrate systematic awareness of the \textsc{uniqueness} and \textsc{novelty} requirement, leading to 75\% accuracy on \texttt{p(sg|pos\_neg)>p(sg|pos\_pos)}, where no models except for Llama 3 70B achieve greater than chance performance. 

\begin{figure*}[h]
\centering
\includegraphics[scale=0.78]{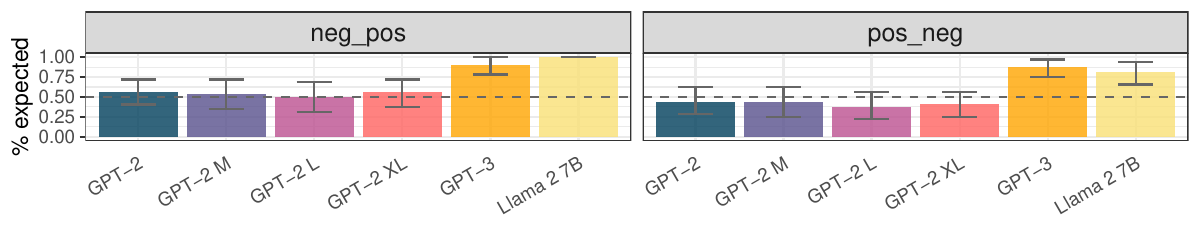}
\caption{Decomposition of results for \textit{affirmative-negation} type sentences in \citet{schuster-linzen-2022-sentence} by \textsc{distance}. Data for GPT-2, GPT-2 M, GPT-2 L, GPT-2 XL, and GPT-3 are retrieved from their GitHub Repository.}
\label{fig:exp1schu_distance_an}
\end{figure*}

The presence of this distance effect provides a possible explanation of the failure in systematic behavior that SL observed in their work. A reanalysis of their data (which involved only \texttt{pos\_neg} and \texttt{neg\_pos} cases) finds that the position of the DE-introducing sentence impacts model accuracy (Figure \ref{fig:exp1schu_distance_an}; $p = 0.0161$).\footnote{We applied a linear mixed-effect model using the \texttt{lme4} \citep{lme4cite} library in R with a main effect of \textsc{distance} and a random effect of \textsc{item}, collapsing across different models. We also examined the effect of \textsc{distance} with respect to other sentence types, where \textsc{distance} is also significant. \haveappendix{See Appendix \ref{sec:moresys} for the corresponding plots and significance testing.}
}

\subsubsection{Plural Continuations} \label{subsubsec:plu}
Plural continuations are felicitous only following \texttt{pos\_pos} contexts. Hence, we would expect the probability of a plural continuation given \texttt{pos\_pos} to be greater than those given \texttt{pos\_neg}, \texttt{neg\_pos}, and \texttt{neg\_neg}. 

\begin{figure*}[h]
    \centering
    \includegraphics[scale=0.78]{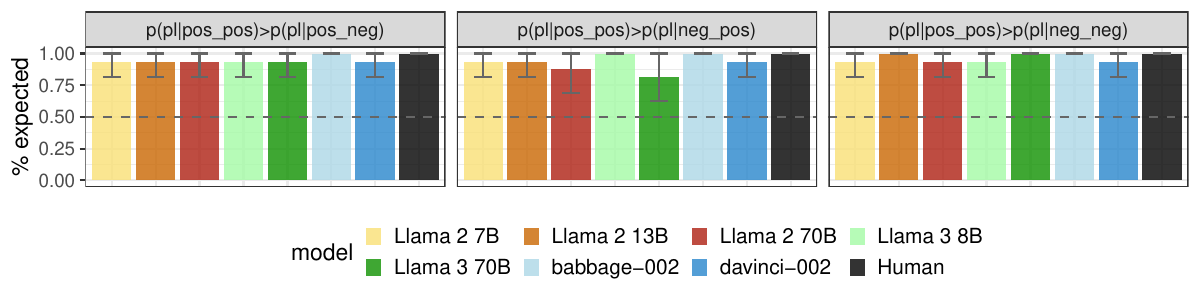}
    \caption{Results for plural comparisons by model and comparison type.}
    \label{fig:exp1plural}
\end{figure*}

Results for these three comparisons are shown in Figure \ref{fig:exp1plural}. All models (and humans) exhibit near-ceiling accuracy, which demonstrates that out of the four possible contexts, they consider \texttt{pos\_pos} to be the best one prior to plural continuations, consistent with human judgments. The lower accuracy of Llama 3 70B compared to all other models again supports the observation from Figure \ref{fig:exp1sing} that larger models do not always perform better than smaller ones in terms of the properties we identified in LIEDER. Regardless, the models' ceiling performance on plural continuations serves as evidence that they have learned both the existence and the plurality requirements for plural definite descriptions.

\subsubsection{Comparing Singluar and Plural Continuations} \label{subsubsec:singandplur}
We finally compare across singular and plural continuations. There are 8 such comparison types in total.
We will focus on three of them that are particularly informative, which are shown in Figure \ref{fig:exp1singularplural}. 
\haveappendix{See Appendix \ref{subsec:addplots1} for results of all comparisons.}

\begin{figure*}[h]
    \centering
    \includegraphics[scale=0.78]{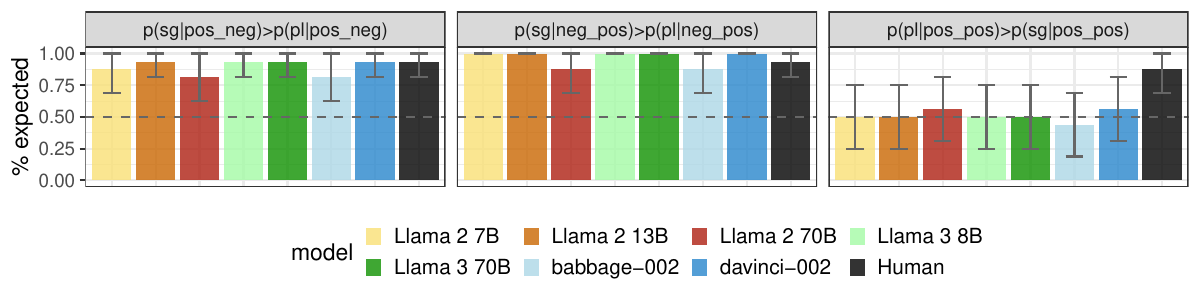}
    \caption{Results by model and comparison type for comparisons across singular and plural continuations.}
    \label{fig:exp1singularplural}
\end{figure*}

For  \texttt{p(sg|neg\_pos)>p(pl|neg\_pos)} and \texttt{p(sg|pos\_neg)>p(pl|pos\_neg)}, all of the LLMs achieve near-ceiling accuracy, consistent with human judgments. The high performance on these two comparisons suggests a preference for singular continuations over plural ones when one and only one relevant entity has been introduced into the discourse using a singular definite description. This preference reflects models' knowledge of \textsc{uniqueness}, which speaks against \textbf{Hypothesis 1} that we proposed in Section \ref{subsubsec:sing}. If the models know \textsc{uniqueness}, why do they perform at chance for the comparison \texttt{p(sg|pos\_neg)>p(sg|pos\_pos)} in Figure \ref{fig:exp1sing}? The only possibility left is \textbf{Hypothesis 2}: they take a singular continuation following a \texttt{pos\_pos} context to be possible because they do not recognize that two distinct DEs have been introduced.

The \texttt{p(pl|pos\_pos)>p(sg|pos\_pos)} comparison provides further support for \texttt{pos\_pos} being a difficult context for the models. 
We saw above that all models consider \texttt{pos\_pos} to be the best context preceding plural continuations. Hence, the problem must be that somehow the models assign an incorrectly high probability to \texttt{p(sg|pos\_pos)} because of failure to enforce the \textsc{novelty} condition with two indefinites. We test this hypothesis further in Experiment 2\haveappendix{ (and Experiment 3 in the Appendix)}.

\section{Experiment 2: Facilitating Novelty} \label{sec:exp2}
As described in Section \ref{sec:dataset}, \texttt{pos\_pos} contexts introduce two DEs using two instances of the same indefinite description. For example, in the context sentence \textit{John owns a dog and Mark owns a dog}, two dogs have been introduced using the same indefinite description \textit{a dog}. However, language models might have difficulty understanding that these are two different dogs because not only do they need to recognize DEs through distinct occurrences of the indefinite description \textit{a dog}, but they must also know the \textsc{novelty} condition to consider these distinct occurrences as distinct DEs.

\subsection{Dataset}
One way to test if  LLMs fail to recognize two different DEs in \texttt{pos\_pos} contexts is to use lexical cues that make explicit the distinctness of the first and second entities.  
If performance relative to \texttt{pos\_pos} contexts increases when the distinction is explicit, then there is evidence that the LLMs fail to recognize the distinction in the implicit case, where the presence of multiple DEs results from the \textsc{novelty} condition on indefinites alone. Accordingly, we make the following modification to our dataset: for each context of the type \texttt{pos\_pos}, we add the adjective “different” to the second indefinite description: 
\pex[aboveexskip=3pt, belowexskip=3pt]<different>
\a \textit{Implicit}: John owns a dog and Mark owns a dog.
\vspace{-3pt}
\a \textit{Explicit Novelty}: John owns a dog and Mark owns a different dog.
\xe
Other contexts and continuations are kept the same. 

\subsection{Results}
\begin{figure*}[h]
    \centering
    \includegraphics[scale=0.78]{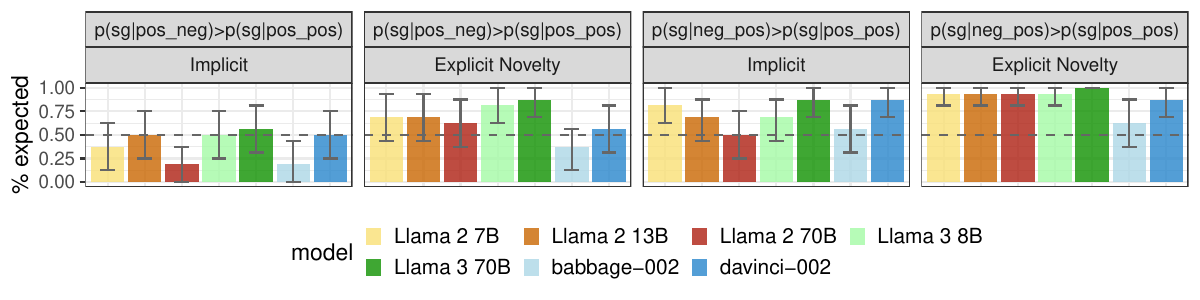}
    \caption{Experiment 2 results by model, version, and comparison type for singular continuations.}
    \label{fig:exp2singular}
\end{figure*}

Results for the felicity comparisons are shown in Figure \ref{fig:exp2singular}.
To quantify the effects of our manipulations, we fit a linear mixed-effect model with two fixed effect predictors: \textsc{version} (a categorical variable with two levels: \textit{Implicit} or \textit{Explicit Novelty}) and \textsc{comparison type}, and \textsc{item} as a random effect. This model gives a significant coefficient for \textsc{version} ($p<0.001$) indicating that accuracy increases significantly from \textit{Implicit} to \textit{Explicit Novelty}. This increase supports \textbf{Hypothesis 2}, i.e., difficulty with the \textsc{novelty} condition. When the distinctness of the two indefinites is made lexically explicit, the models can better recognize that two distinct DEs are introduced in \texttt{pos\_pos} contexts. 
\haveappendix{In Appendix \ref{sec:exp3}, we show that another way of cueing the creation of multiple DEs by explicitly supplying information about plurality achieves the same effect.}

\section{Discussion}
In Experiment 1, we saw that all LLMs displayed clear knowledge of \textsc{existence} and \textsc{plurality}. They inarguably consider singular continuations to be bad after contexts that introduce no DEs and plural continuations to be good only when more than one DE has been introduced. However, they failed to show clear knowledge of \textsc{uniqueness}, as indicated by their at-chance performance for the comparison type \texttt{p(sg|pos\_neg)>p(sg|pos\_pos)}. We believe that such underperformance is not due to a lack of knowledge of \textsc{uniqueness} because the results in Figure \ref{fig:exp1singularplural} support knowledge of the \textsc{uniqueness} requirement. Instead, we believe that the problem lies in \textsc{novelty}. Specifically, the models cannot identify that \texttt{pos\_pos} contexts introduce two distinct DEs. 

We suspect that the \texttt{pos\_pos} context might be particularly challenging because the two DEs are introduced using identical indefinite NPs (i.e., the two occurrences of \textit{a dog} in \textit{John owns a dog and Mark owns a dog too}). One might imagine that it is often sufficient to  associate unique NPs with distinct DEs. For example, given the sentence \textit{John owns a cat and a dog}, the information that \textit{a cat} and \textit{a dog} are two distinct DEs is lexically encoded. If a model adopts such a lexically dependent strategy for DE introduction, then the \texttt{pos\_pos} cases from Experiment 1 are exactly the expected point of failure, as such cases require sensitivity to \textsc{novelty} in order to succeed. Indeed, as shown in Experiment 2, supplying explicit information about distinctiveness does improve model performance.

Taken together, these results all point to language models' mastery of \textsc{existence}, \textsc{uniquness}, and \textsc{plurality} and a lack of knowledge on \textsc{novelty} when minimal information is given.

\paragraph{Distance Effect} Results from Experiment 1 also show a clear effect of \textsc{distance}. In our reanalysis of SL, LMs did a better job of recognizing DEs when the entities are introduced closer to the continuation that refers to them. Within results from our own evaluation suite, there is also a clear preference for introducing DEs closer to the definite description when the singular continuations are equally felicitous. It seems as if as the sentence unfolds, the status of a DE as one that can be referred to gradually decreases. In other words, there is a greater cost associated with referring to a DE that is introduced earlier in time than those that are introduced later.

\section{Related Work} 

The current paper contributes to the body of literature that examines the semantic knowledge of neural language models. \citet{kim-etal-2019-probing} assessed LLM comprehension of functional words, including a subtask of differentiating indefinite and definite determiners. Their stimuli were constructed by swapping \textit{a/an} with \textit{the} in a sentence. Such manipulation implicitly encodes the properties identified in LIEDER: using \textit{the} in the place of \textit{a} often violates \textsc{uniqueness} and \textsc{existence}, and \textsc{novelty} could also be violated vice versa.  

On the discourse level, aside from \citet{schuster-linzen-2022-sentence} from which the current paper draws inspiration, \citet{upadhye-etal-2020-predicting} examined contexts with different biases concerning the DE that will be mentioned next. They found that unlike humans, GPT-2 \citep{radford2019language} and Transformer-XL \citep{dai-etal-2019-transformer} are not sensitive to the manipulation of contexts when predicting the entity that will be mentioned next. \citet{loaiciga-etal-2022-new} built probing models to investigate whether pretrained representations encode information about an entity being newly introduced or having been mentioned before (which they call discourse-new/-old respectively). They found that while a high F1 score can be achieved for the classification of discourse-new vs.\ discourse-old, models struggle to locate the entities within a sequence, a finding that is in line with our result that models lack understanding of the \textsc{novelty} requirement. Lastly, \citet{kim-schuster-2023-entity} studied LLM ability to accurately represent the states of discourse entities across long narratives and observed that only LLMs that have been pretrained on code exhibited non-trivial entity tracking abilities.

The LIEDER dataset also adds to the body of work that uses linguistic insights in the development of semantic benchmarks for LLMs.  This work includes COGS \citep{kim-linzen-2020-cogs}, ReCOGS \citep{recogs}, and SLOG \citep{li-etal-2023-slog} on compositional generalization and \textsc{ImpPres} \citep{jeretic-etal-2020-natural}, NOPE \citep{parrish-etal-2021-nope}, \citet{kim-etal-2021-linguist}, and (QA)$^2$ \citep{kim-etal-2023-qa} that assess models' ability in handling presuppositions. 

\section{Conclusion}
In this paper, we proposed the Linguistically-Informed Evaluation for Discourse Entity Recognition (LIEDER) dataset. Our paradigm allows for a detailed examination of language models' knowledge of four properties of definite and indefinite NPs concerning their ability to introduce and refer to DEs.  These properties are \textsc{existence}, \textsc{uniqueness}, \textsc{plurality}, and \textsc{novelty}. We demonstrated that despite mastering \textsc{existence}, \textsc{uniqueness}, and \textsc{plurality}, the LLMs we examined lack understanding of the \textsc{novelty} requirement.
In spite of this deficiency, we showed that language models of the Llama 2, Llama 3, and GPT series reflected the human preference of referring to DEs that are introduced closer to their reference point, which we label an effect of \textsc{distance}. We recognize that given the fast-paced development of LLM research, it is highly likely that the performance presented in the current paper will be surpassed by future generations of LMs. However, the success of LIEDER in helping to identify language models' deficiency in DE recognition highlights the importance of linguistic considerations in evaluating the strengths and weaknesses of future language models.

\section*{Ethics Statement}
\paragraph{Limitations}
The current study only focuses on English, which has overt determiners for indefinite and definite NPs. There are other languages that do not have determiners equivalent to \textit{a} and \textit{the} in English. For example, Mandarin Chinese makes use of demonstratives that can serve similar purposes as English determiners. Our evaluation paradigm is thus limited in that it cannot be directly used to evaluate DE recognition on language models trained on other languages without considering language-specific properties relating to DE introduction and reference.

\paragraph{Risks}
All participants in the human experiment were recruited through Prolific under the approval of the Yale University IRB. At the beginning of the experiment, they were presented with consent forms that indicated the potential risks associated with participation, and only those who consented were allowed to proceed with the task. The risk was minimal. All identifier information has been removed from the data to guarantee anonymity. Participants received compensation equivalent to \$12/hr, which is around 70\% higher than the federal minimum wage.

\bibliography{anthology,custom}

\begin{thebibliography}{24}
\expandafter\ifx\csname natexlab\endcsname\relax\def\natexlab#1{#1}\fi

\bibitem[{Bates et~al.(2015)Bates, M{\"a}chler, Bolker, and Walker}]{lme4cite}
Douglas Bates, Martin M{\"a}chler, Ben Bolker, and Steve Walker. 2015.
\newblock \href {https://doi.org/10.18637/jss.v067.i01} {Fitting linear mixed-effects models using {lme4}}.
\newblock \emph{Journal of Statistical Software}, 67(1):1--48.

\bibitem[{Dai et~al.(2019)Dai, Yang, Yang, Carbonell, Le, and Salakhutdinov}]{dai-etal-2019-transformer}
Zihang Dai, Zhilin Yang, Yiming Yang, Jaime Carbonell, Quoc Le, and Ruslan Salakhutdinov. 2019.
\newblock \href {https://doi.org/10.18653/v1/P19-1285} {Transformer-{XL}: Attentive language models beyond a fixed-length context}.
\newblock In \emph{Proceedings of the 57th Annual Meeting of the Association for Computational Linguistics}, pages 2978--2988, Florence, Italy. Association for Computational Linguistics.

\bibitem[{Heim and Kratzer(1998)}]{heim1998semantics}
Irene Heim and Angelika Kratzer. 1998.
\newblock \emph{Semantics in Generative Grammar}.
\newblock Blackwell, Malden, MA.

\bibitem[{Heim(1982)}]{heim1982semantics}
Irene~Roswitha Heim. 1982.
\newblock \emph{The semantics of definite and indefinite noun phrases}.
\newblock University of Massachusetts Amherst.

\bibitem[{Hu and Levy(2023)}]{hu2023prompting}
Jennifer Hu and Roger Levy. 2023.
\newblock Prompting is not a substitute for probability measurements in large language models.
\newblock In \emph{Proceedings of the 2023 Conference on Empirical Methods in Natural Language Processing}, pages 5040--5060.

\bibitem[{Jeretic et~al.(2020)Jeretic, Warstadt, Bhooshan, and Williams}]{jeretic-etal-2020-natural}
Paloma Jeretic, Alex Warstadt, Suvrat Bhooshan, and Adina Williams. 2020.
\newblock \href {https://doi.org/10.18653/v1/2020.acl-main.768} {Are natural language inference models {IMPPRESsive}? {L}earning {IMPlicature} and {PRESupposition}}.
\newblock In \emph{Proceedings of the 58th Annual Meeting of the Association for Computational Linguistics}, pages 8690--8705, Online. Association for Computational Linguistics.

\bibitem[{Kim et~al.(2023)Kim, Htut, Bowman, and Petty}]{kim-etal-2023-qa}
Najoung Kim, Phu~Mon Htut, Samuel~R. Bowman, and Jackson Petty. 2023.
\newblock \href {https://doi.org/10.18653/v1/2023.acl-long.472} {({QA})$^2$: Question answering with questionable assumptions}.
\newblock In \emph{Proceedings of the 61st Annual Meeting of the Association for Computational Linguistics (Volume 1: Long Papers)}, pages 8466--8487, Toronto, Canada. Association for Computational Linguistics.

\bibitem[{Kim and Linzen(2020)}]{kim-linzen-2020-cogs}
Najoung Kim and Tal Linzen. 2020.
\newblock \href {https://doi.org/10.18653/v1/2020.emnlp-main.731} {{COGS}: A compositional generalization challenge based on semantic interpretation}.
\newblock In \emph{Proceedings of the 2020 Conference on Empirical Methods in Natural Language Processing (EMNLP)}, pages 9087--9105, Online. Association for Computational Linguistics.

\bibitem[{Kim et~al.(2019)Kim, Patel, Poliak, Xia, Wang, McCoy, Tenney, Ross, Linzen, Van~Durme, Bowman, and Pavlick}]{kim-etal-2019-probing}
Najoung Kim, Roma Patel, Adam Poliak, Patrick Xia, Alex Wang, Tom McCoy, Ian Tenney, Alexis Ross, Tal Linzen, Benjamin Van~Durme, Samuel~R. Bowman, and Ellie Pavlick. 2019.
\newblock \href {https://doi.org/10.18653/v1/S19-1026} {Probing what different {NLP} tasks teach machines about function word comprehension}.
\newblock In \emph{Proceedings of the Eighth Joint Conference on Lexical and Computational Semantics (*{SEM} 2019)}, pages 235--249, Minneapolis, Minnesota. Association for Computational Linguistics.

\bibitem[{Kim et~al.(2021)Kim, Pavlick, Karagol~Ayan, and Ramachandran}]{kim-etal-2021-linguist}
Najoung Kim, Ellie Pavlick, Burcu Karagol~Ayan, and Deepak Ramachandran. 2021.
\newblock \href {https://doi.org/10.18653/v1/2021.acl-long.304} {Which linguist invented the lightbulb? presupposition verification for question-answering}.
\newblock In \emph{Proceedings of the 59th Annual Meeting of the Association for Computational Linguistics and the 11th International Joint Conference on Natural Language Processing (Volume 1: Long Papers)}, pages 3932--3945, Online. Association for Computational Linguistics.

\bibitem[{Kim and Schuster(2023)}]{kim-schuster-2023-entity}
Najoung Kim and Sebastian Schuster. 2023.
\newblock \href {https://doi.org/10.18653/v1/2023.acl-long.213} {Entity tracking in language models}.
\newblock In \emph{Proceedings of the 61st Annual Meeting of the Association for Computational Linguistics (Volume 1: Long Papers)}, pages 3835--3855, Toronto, Canada. Association for Computational Linguistics.

\bibitem[{Landman(1989)}]{landman1989groups}
Fred Landman. 1989.
\newblock Groups, i.
\newblock \emph{Linguistics and philosophy}, pages 559--605.

\bibitem[{Li et~al.(2023)Li, Donatelli, Koller, Linzen, Yao, and Kim}]{li-etal-2023-slog}
Bingzhi Li, Lucia Donatelli, Alexander Koller, Tal Linzen, Yuekun Yao, and Najoung Kim. 2023.
\newblock \href {https://doi.org/10.18653/v1/2023.emnlp-main.194} {{SLOG}: A structural generalization benchmark for semantic parsing}.
\newblock In \emph{Proceedings of the 2023 Conference on Empirical Methods in Natural Language Processing}, pages 3213--3232, Singapore. Association for Computational Linguistics.

\bibitem[{Li et~al.(2022)Li, Sun, Han, and Li}]{nersurvey}
Jing Li, Aixin Sun, Jianglei Han, and Chenliang Li. 2022.
\newblock \href {https://doi.org/10.1109/TKDE.2020.2981314} {A survey on deep learning for named entity recognition}.
\newblock \emph{IEEE Transactions on Knowledge and Data Engineering}, 34(1):50--70.

\bibitem[{Lo{\'a}iciga et~al.(2022)Lo{\'a}iciga, Beyer, and Schlangen}]{loaiciga-etal-2022-new}
Sharid Lo{\'a}iciga, Anne Beyer, and David Schlangen. 2022.
\newblock \href {https://aclanthology.org/2022.coling-1.73} {New or old? exploring how pre-trained language models represent discourse entities}.
\newblock In \emph{Proceedings of the 29th International Conference on Computational Linguistics}, pages 875--886, Gyeongju, Republic of Korea. International Committee on Computational Linguistics.

\bibitem[{{Meta AI}(2024)}]{llama3}
{Meta AI}. 2024.
\newblock \href {https://ai.meta.com/blog/meta-llama-3/} {Introducing {Meta} {Llama} 3: {The} most capable openly available {LLM} to date}.

\bibitem[{Parrish et~al.(2021)Parrish, Schuster, Warstadt, Agha, Lee, Zhao, Bowman, and Linzen}]{parrish-etal-2021-nope}
Alicia Parrish, Sebastian Schuster, Alex Warstadt, Omar Agha, Soo-Hwan Lee, Zhuoye Zhao, Samuel~R. Bowman, and Tal Linzen. 2021.
\newblock \href {https://doi.org/10.18653/v1/2021.conll-1.28} {{NOPE}: A corpus of naturally-occurring presuppositions in {E}nglish}.
\newblock In \emph{Proceedings of the 25th Conference on Computational Natural Language Learning}, pages 349--366, Online. Association for Computational Linguistics.

\bibitem[{Radford et~al.(2019)Radford, Wu, Child, Luan, Amodei, Sutskever et~al.}]{radford2019language}
Alec Radford, Jeffrey Wu, Rewon Child, David Luan, Dario Amodei, Ilya Sutskever, et~al. 2019.
\newblock Language models are unsupervised multitask learners.
\newblock \emph{OpenAI blog}, 1(8):9.

\bibitem[{Russell(1905)}]{russell1905denoting}
Bertrand Russell. 1905.
\newblock On denoting.
\newblock \emph{Mind}, 14:479--493.

\bibitem[{Schuster and Linzen(2022)}]{schuster-linzen-2022-sentence}
Sebastian Schuster and Tal Linzen. 2022.
\newblock \href {https://doi.org/10.18653/v1/2022.naacl-main.71} {When a sentence does not introduce a discourse entity, transformer-based models still sometimes refer to it}.
\newblock In \emph{Proceedings of the 2022 Conference of the North American Chapter of the Association for Computational Linguistics: Human Language Technologies}, pages 969--982, Seattle, United States. Association for Computational Linguistics.

\bibitem[{Touvron et~al.(2023)Touvron, Martin, Stone, Albert, Almahairi, Babaei, Bashlykov, Batra, Bhargava, Bhosale et~al.}]{touvron2023llama2}
Hugo Touvron, Louis Martin, Kevin Stone, Peter Albert, Amjad Almahairi, Yasmine Babaei, Nikolay Bashlykov, Soumya Batra, Prajjwal Bhargava, Shruti Bhosale, et~al. 2023.
\newblock Llama 2: Open foundation and fine-tuned chat models.
\newblock \emph{arXiv preprint arXiv:2307.09288}.

\bibitem[{Upadhye et~al.(2020)Upadhye, Bergen, and Kehler}]{upadhye-etal-2020-predicting}
Shiva Upadhye, Leon Bergen, and Andrew Kehler. 2020.
\newblock \href {https://doi.org/10.18653/v1/2020.emnlp-main.70} {Predicting reference: What do language models learn about discourse models?}
\newblock In \emph{Proceedings of the 2020 Conference on Empirical Methods in Natural Language Processing (EMNLP)}, pages 977--982, Online. Association for Computational Linguistics.

\bibitem[{Wolf et~al.(2019)Wolf, Debut, Sanh, Chaumond, Delangue, Moi, Cistac, Rault, Louf, Funtowicz et~al.}]{wolf2019huggingface}
Thomas Wolf, Lysandre Debut, Victor Sanh, Julien Chaumond, Clement Delangue, Anthony Moi, Pierric Cistac, Tim Rault, R{\'e}mi Louf, Morgan Funtowicz, et~al. 2019.
\newblock Huggingface's transformers: State-of-the-art natural language processing.
\newblock \emph{arXiv preprint arXiv:1910.03771}.

\bibitem[{Wu et~al.(2023)Wu, Manning, and Potts}]{recogs}
Zhengxuan Wu, Christopher~D. Manning, and Christopher Potts. 2023.
\newblock \href {https://doi.org/10.1162/tacl_a_00623} {{{ReCOGS}: How Incidental Details of a Logical Form Overshadow an Evaluation of Semantic Interpretation}}.
\newblock \emph{Transactions of the Association for Computational Linguistics}, 11:1719--1733.

\end{thebibliography}

\haveappendix{

\appendix

\section{Human Experiments} \label{sec:morehuman}
A total of 80 participants were recruited through Prolific under the approval of the Yale University IRB. Screener conditions were set such that all participants were native English speakers with perfect or corrected vision and without language disorders.

We used the Gorilla experimental platform to present a context sentence for a period of 300ms per word, and then presented the continuation sentence, again for 300ms per word. Participants then moved the dot on the slider scale appearing below to indicate their judgment. Figure \ref{fig:humaninterface} demonstrates this experimental interface.

\begin{figure*}
    \centering
    \includegraphics[scale=0.3]{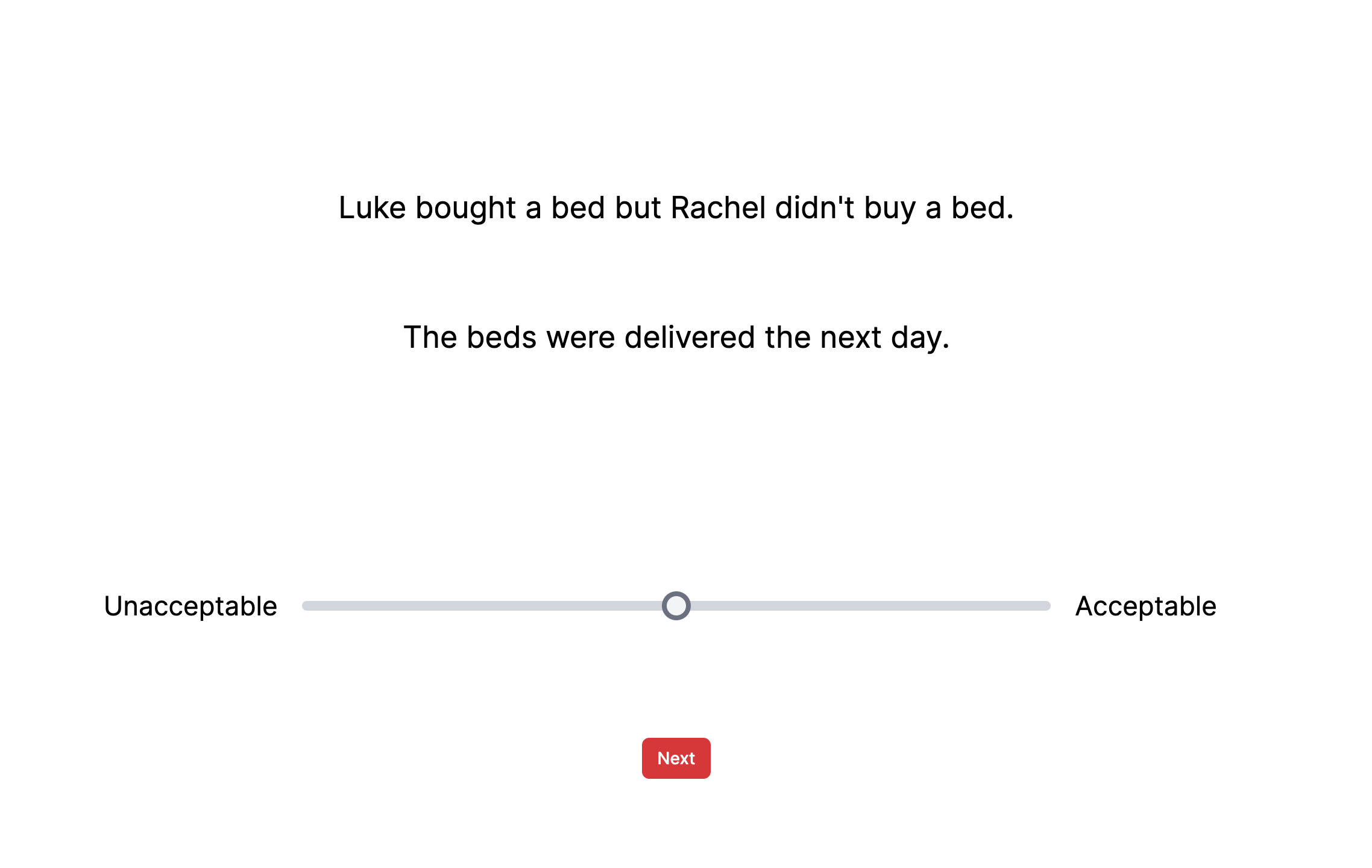}
    \caption{Experimental Interface on Gorilla. The first sentence on the screen is a target item of the \texttt{pos\_neg} category Since the continuation is plural, it is expected to be unacceptable if following the context.}
    \label{fig:humaninterface}
\end{figure*}

Each participant received 26 trials in total which were composed of 2 practice trials and 16 target trials with 8 control trials that appeared in random order. In the practice trials, they were instructed to move the slider all the way to the right if they thought the continuation was perfectly acceptable, and all the way to the left if clearly unacceptable. In the 16 target trials, they were exposed to the same set of stimuli as the language models corresponding to 16 different sentence frames (i.e. \textsc{item}). The filler trials were the same for each participant where the judgments for each of them were either strictly felicitous or infelicitous. Each participant received compensation that averaged \$12.20/hr.

Among the initial 80 participants we recruited, data from four participants was excluded because they failed to answer 7 of the 8 control items correctly.


\section{Supplementary Plots for Experiment 1 and 2} \label{subsec:addplots1}

All results reported in the experiments were concerning \textit{affirmative-negation} sentences. SL identify four pairs of operators where the operators in each pair differ in the way they modulate DE introduction. They are \textit{affirmative-negation}, \textit{affirmative-modal}, \textit{know-doubt}, and \textit{managed-failed}, which is exemplified in Table \ref{tab:operator}. The full LIEDER dataset includes conjoined sentences of all of these types except for \textit{affirmative-modal}. This decision is based on our judgment that it is easier to get a wide-scope reading of indefinites when they are embedded in modals than when they are embedded in negations.
\ex<ambad>
John wants to own a dog and Mark owns a dog. The dog is cute.
\xe
For example, according to the design in LIEDER, (\getref{ambad}) is of type \texttt{neg\_pos}, and it is intended that the second conjunct introduces a discourse entity but not the first one. However, it is easy to get the reading that there is a specific dog that Mark wants to own, thus making the singular definite infelicitous. To avoid complexities like this, we decided to focus our analysis on the other three contrast types instead.

\begin{table*}[] 
    \centering
    \begin{tabular}{ccc}
    \toprule
Operator Type &     \texttt{pos}   &  \texttt{neg} \\ \midrule
 \textit{affirmative-negation} &    John owns a dog.     & John doesn't own a dog.\\
 \textit{affirmative-modal} & John owns a dog. & John wants to own a dog.\\
 \textit{know-doubt} & I know that John owns a dog. & I doubt that John owns a dog. \\
 \textit{managed-failed} & John managed to adopt a dog. & John failed to adopt a dog.\\
 \bottomrule
    \end{tabular}
    \caption{Four pairs of sentential operators introduced by SL. The \texttt{pos} column indicates cases where DEs are introduced. The \texttt{neg} column indicates cases where DEs are not introduced. All operator types are included in LIEDER except for \textit{affirmative-modal}.}
    \label{tab:operator}
\end{table*}

In the rest of this section, we show results for two other types of sentences: \textit{know-doubt} and \textit{managed-failed}. Note that we did not elicit human judgments for \textit{know-doubt} sentences as some of these sentences are too long to format on the Gorilla interface.

\subsection{Experiment 1}
Figure \ref{fig:moreexp1_singular} corresponds to singular continuations, Figure \ref{fig:moreexp1_plural} plural continuations, and Figure \ref{fig:moreexp1_sp} for singular and plural comparisons.

\begin{figure*}[h]
\centering
\includegraphics[scale=0.6]{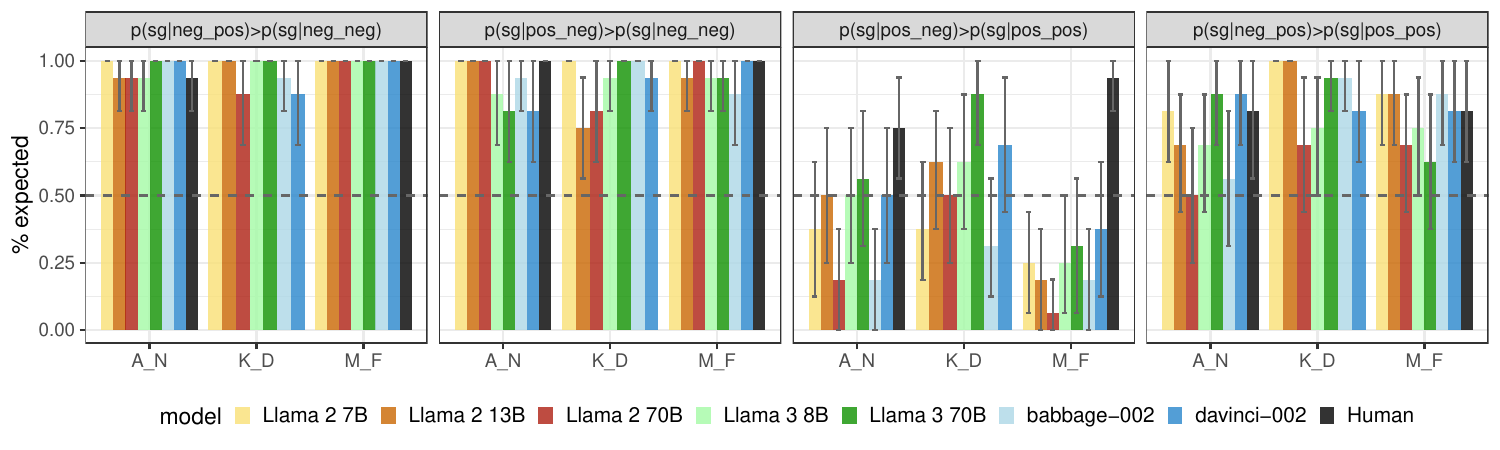}
\caption{Results of all three sentence types from Experiment 1 - Singular Continuations. A\_N, K\_D, M\_F stand for \textit{affirmative-negation}, \textit{know-doubt}, and \textit{managed-failed} respectively.}
\label{fig:moreexp1_singular}
\end{figure*}

\begin{figure*}[h]
\centering
\includegraphics[scale=0.6]{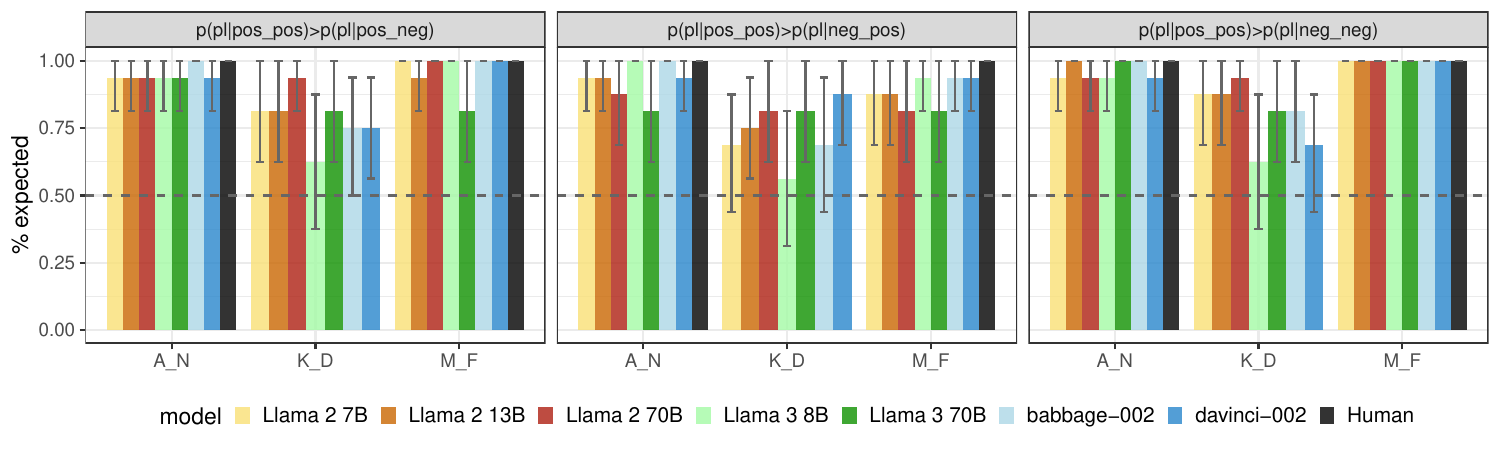}
\caption{Results of all three sentence types from Experiment 1 - Plural Continuations.}
\label{fig:moreexp1_plural}
\end{figure*}

\begin{figure*}[h]
\centering
\includegraphics[scale=0.6]{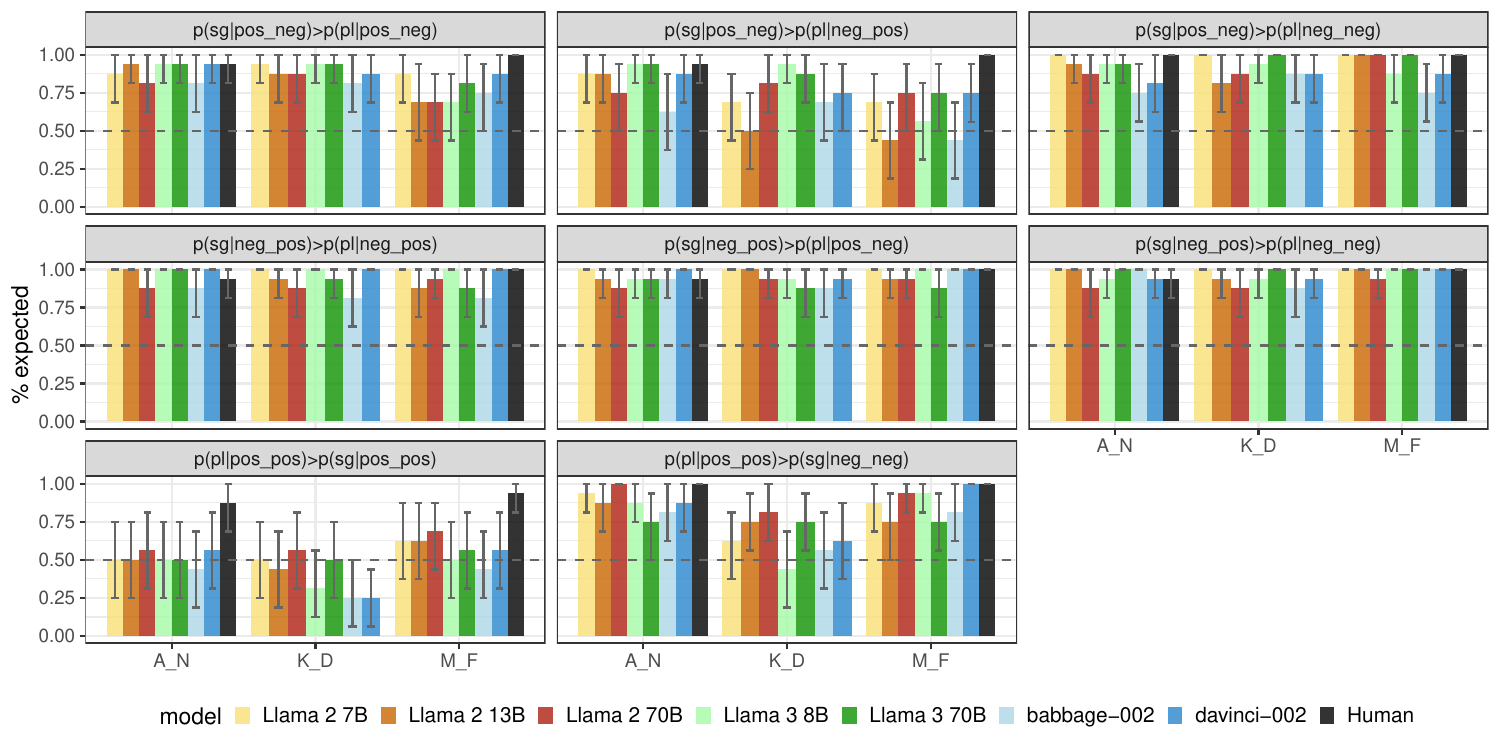}
\caption{Results of all three sentence types from Experiment 1 - Singular and Plural Comparisons.}
\label{fig:moreexp1_sp}
\end{figure*}

\subsection{Experiment 2}
Figure \ref{fig:moreexp2} shows the increase from \textit{Implicit} to \textit{Explicit Novelty} for all three types of sentences. The accuracy increase from \textit{Implicit} to \textit{Explicit Novelty} is significant ($p<0.001$) under the same linear mixed-effect model specified in Section \ref{sec:exp2} that collapses across language models and sentence type.

\begin{figure*}[h]
\centering
\includegraphics[scale=0.6]{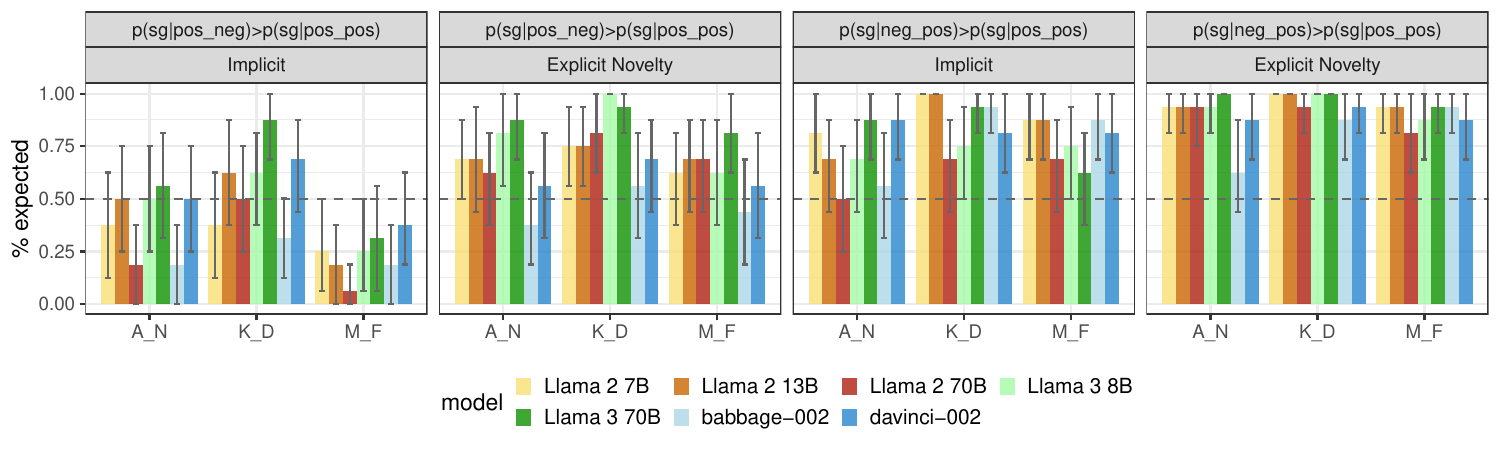}
\caption{Results of all three sentence types from Experiment 2 - Implicit vs. Explicit Novelty.}
\label{fig:moreexp2}
\end{figure*}

\section{Effects of \textsc{distance} in SL} \label{sec:moresys}

In Section \ref{subsubsec:sing}, we showed the effect of \textsc{distance} on \textit{affirmative-negation} sentences. Here in Figure \ref{fig:exp1schu_distance}, we provide a comprehensive plot showing the effect of distance on all four types of sentences. The same linear mixed-effect model was applied. The main effect of \textsc{distance} is still significant ($p<0.001$). 

\begin{figure*}[h]
\centering
\includegraphics[scale=0.6]{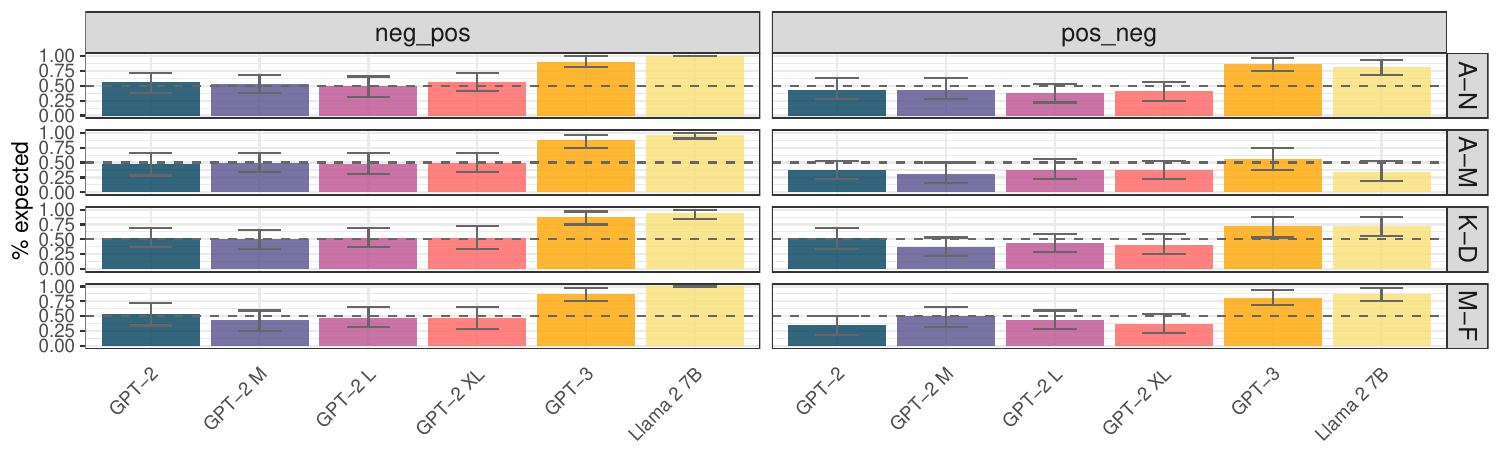}
\caption{Decomposition of results in SL by distance and sentence type.}
\label{fig:exp1schu_distance}
\end{figure*}

\section{Experiment 3: Plural Indefinites} \label{sec:exp3}
We conducted a third experiment, where we introduced a third type of conjunct besides \texttt{pos} and \texttt{neg} which we call \texttt{two}. In the \texttt{two} conjunct, a plural indefinite is used as an explicit cue that there is more than one relevant entity in the discourse. Consider the following distinction:

\pex<two>
\a \textit{Implicit}: John owns a dog and Mark owns a dog too. (\texttt{pos\_pos)}
\a \textit{Explicit Plurality}: John owns two dogs and Mark doesn't own a dog. (\texttt{two\_neg})
\xe

Both (\getref{two}a) and (\getref{two}b) involve the introduction of two dogs into the discourse. However, in (\getref{two}a), the \textsc{novelty} condition is necessary to conclude the existence of two distinct dogs, one owned by John and the other owned by Mark. In (\getref{two}b), the fact that there are two dogs is directly encoded in the phrase \textit{two dogs}. Hence, if our hypothesis about the models' difficulties with the \texttt{pos\_pos} condition is correct, this way of directly supplying information about plurality in this way will increase models' preference for singular definites in contexts where only one DE as compared to contexts in which multiple discourse references are introduced. In other words, we expect there to be an increase in accuracy from \texttt{p(sg|pos\_neg)>p(sg|pos\_pos)} comparisons
 to \texttt{p(sg|pos\_neg)>p(sg|two\_neg)}, and similarly from \texttt{p(sg|neg\_pos)>p(sg|pos\_pos)} to \texttt{p(sg|neg\_pos)>p(sg|neg\_two)}.

\subsection{Dataset}
We make the following changes to the \textit{Implicit} dataset in Experiment 1, resulting in a new dataset that we will call \textit{Explicit Plurality} henceforth. For each sentence where exactly one DE is introduced (i.e., \texttt{pos\_neg} and \texttt{neg\_pos}), we add another one where the conjunct of type \texttt{two} replaces \texttt{pos}. This results in four more context-continuation combinations, given in Table \ref{tab:twofelicity}, of which two are felicitous and two are infelicitous. 

\begin{table}[] 
\centering
\begin{tabular}{cc}
\toprule
New Combinations         & Felicity     \\
\midrule
\texttt{p(sg|two\_neg)} & infelicitous \\
\texttt{p(sg|neg\_two)} & infelicitous \\
\texttt{p(pl|two\_neg)} & felicitous  \\
\texttt{p(pl|neg\_two)} & felicitous \\
\bottomrule
\end{tabular}
\caption{Additional context-continuation combinations in Experiment 3 and their corresponding felicity judgments.}
\label{tab:twofelicity}
\end{table}

\subsection{Results}
Figure \ref{fig:exp3singular} shows the results of comparisons of the probability of singular continuations in contexts that should evoke a single discourse referent as compared to those that should evoke two. Columns 1 and 3 are of the category \textit{Implicit}, whereas columns 2 and 4 belong to \textit{Explicit Plurality}. All of the LLMs show a clearer dispreference for contexts that evoke multiple discourse referents when such evocation is done by a single plural indefinite as compared to two singular indefinites. Model failure to recognize two distinct DEs from \texttt{pos\_pos} is again supported by experimental data.
\begin{figure*}[h]
    \centering
    \includegraphics[scale=0.6]{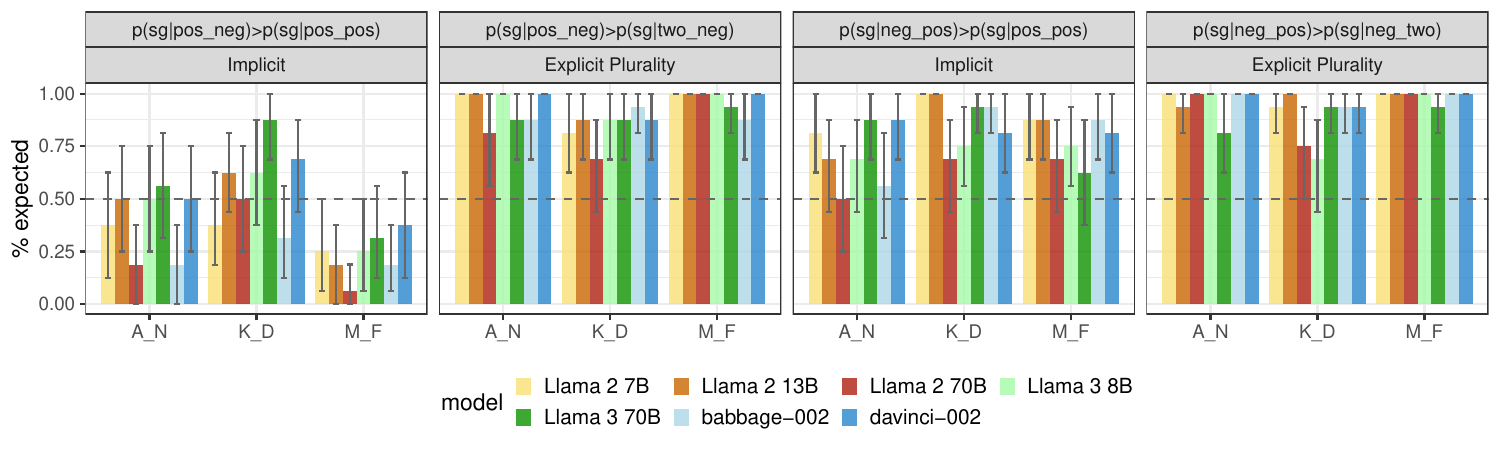}
    \caption{Results of all three sentence types from Experiment 3 - Implicit vs. Explicit Plurality.}
    \label{fig:exp3singular}
\end{figure*}

\section{Results Under A Different Metric}
In Section \ref{subsubsec:sing} and \ref{subsubsec:plu}, we compared the probabilities that the models assign to the same continuation given two different contexts, one felicitous and the other infelicitous. There are two ways to operationalize such comparisons according to SL. 

The first one is to use a direct metric which we adopted in all of the plots presented above. 
\pex <metric2>
    \a \textsc{F:} John owns a dog but Mark doesn't own a dog.
    \a \textsc{I:} John owns a dog and Mark owns a dog as well.\\
    \textsc{Target:} The dog is very cute.
\xe
Using (\getref{metric2}) as an example, we expect the following inequality to hold: 

\begin{equation} \label{eq:exp2metric}
p(\textsc{target}|\textsc{F})>p(\textsc{target}|\textsc{I}). 
\end{equation}
Thus, we can compute accuracy for a given continuation with respect to a pair of environments, one felicitous and the other infelicitous, by measuring the proportion of times this inequality is satisfied.

However, as SL note in their Experiment 1, this metric can be problematic given that the two probabilities in the inequality are essentially drawn from different distributions, so it is possible that the probabilities are underestimated -- if the language model considers that, say, given the context of \textsc{F} (\texttt{pos\_neg}) in (\getref{metric2}), there is some other continuation that is highly likely. Thus, the probability of \textit{The dog is very cute} given this context can be smaller than its infelicitous \texttt{pos\_pos} counterpart, although the language model may consider \texttt{pos\_pos} to be less acceptable.

To solve this issue, \citet{schuster-linzen-2022-sentence} proposed a second metric which introduces control examples involving the non-coreferential continuation such as the following.
\ex<nonref>
\textsc{ContNonref}: It's not a big deal.
\xe
Using \textsc{ContNonref}, we now compare two fractions (\ref{eq1}) and (\ref{eq2}). Specifically, (\ref{eq1}) is expected to be greater than (\ref{eq2}).

\begin{equation} \label{eq1}
\frac{p(\textsc{Cont}|\textsc{F})}{p(\textsc{Cont}|\textsc{F})+ p(\textsc{ContNonref}|\textsc{F})}
\end{equation}
\begin{equation} \label{eq2}
    \frac{p(\textsc{Cont}|\textsc{I})}{p(\textsc{Cont}|\textsc{I})+ p(\textsc{ContNonref}|\textsc{I})}
\end{equation}

Results for singular, plural, and singular vs. plural continuations using the relative metric are shown in Figure \ref{fig:appendix_nonref_sing}, Figure \ref{fig:appendix_nonref_plu}, and Figure \ref{fig:appendix_nonref_sp} respectively. These results are qualitatively the same as the ones under the direct metric that we presented in the main body of the paper.

\begin{figure*}[h] 
\centering
\includegraphics[scale=0.6]{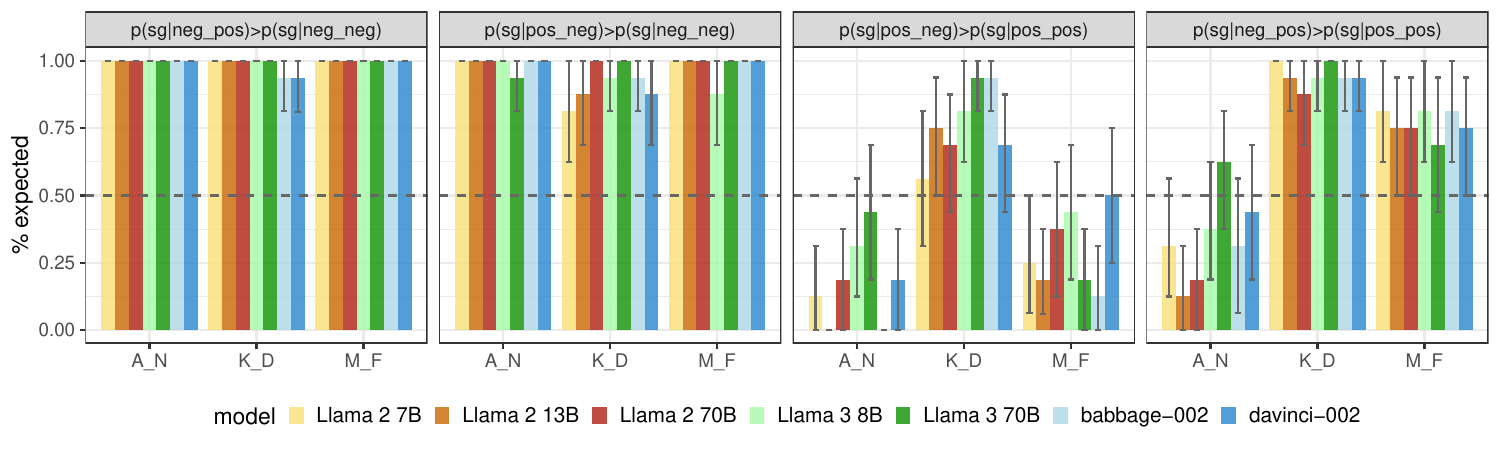}
\caption{Results for singular continuations by model and comparison type under the relative metric.
}
\label{fig:appendix_nonref_sing}
\end{figure*}

\begin{figure*}[h] 
\centering
\includegraphics[scale=0.6]{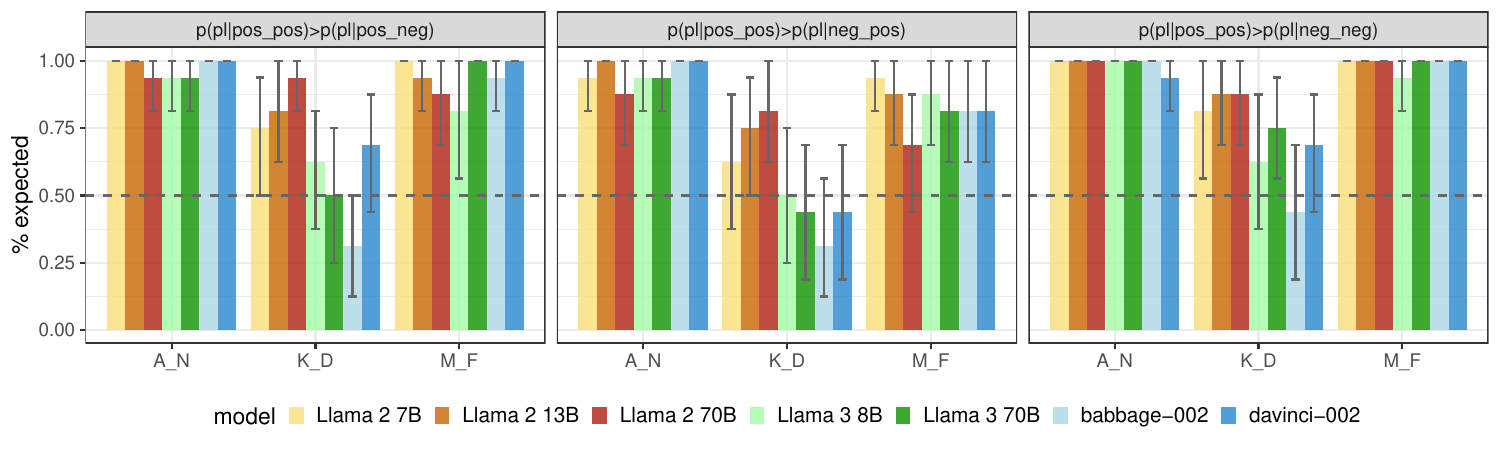}
\caption{Results for plural continuations by model and comparison type under the relative metric.
}
\label{fig:appendix_nonref_plu}
\end{figure*}

\begin{figure*}[h] 
\centering
\includegraphics[scale=0.6]{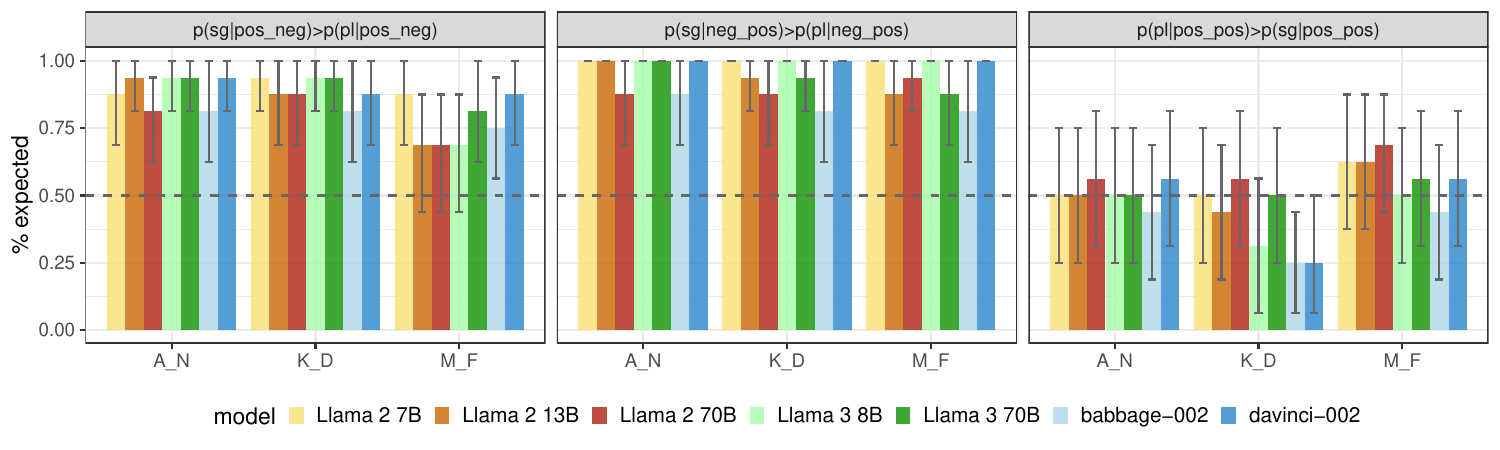}
\caption{Results for comparisons across singular and plural continuations by model and comparison type under the relative metric.
}
\label{fig:appendix_nonref_sp}
\end{figure*}

}

\end{document}